\documentclass[letterpaper, 10 pt, conference]{ieeeconf}  

\IEEEoverridecommandlockouts                              

\usepackage[T1]{fontenc}
\usepackage{amsmath}
\usepackage{amssymb}
\usepackage{graphicx}
\usepackage{booktabs}
\title{\LARGE \bf
Leveraging Stable Diffusion for Monocular Depth Estimation via Image Semantic Encoding
}

\author{
Jingming Xia, Guanqun Cao, Guang Ma$^{*}$, Yiben Luo$^{*}$, Qinzhao Li$^{*}$, John Oyekan$^{\dagger}$
\thanks{All authors are with the University of York, York, UK}
\thanks{$^{*}$Refers to equal contribution}
\thanks{$^{\dagger}$Refers to corresponding author}
}

\begin{document}

\maketitle
\thispagestyle{empty}
\pagestyle{empty}

\begin{abstract}

Monocular depth estimation involves predicting depth from a single RGB image and plays a crucial role in applications such as autonomous driving, robotic navigation, 3D reconstruction, etc. Recent advancements in learning-based methods have significantly improved depth estimation performance. Generative models, particularly Stable Diffusion, have shown remarkable potential in recovering fine details and reconstructing missing regions through large-scale training on diverse datasets. However, models like CLIP, which rely on textual embeddings, face limitations in complex outdoor environments where rich context information is needed. These limitations reduce their effectiveness in such challenging scenarios. Here, we propose  a novel image-based semantic embedding that extracts contextual information directly from visual features, significantly improving depth prediction in complex environments. Evaluated on the KITTI and Waymo datasets, our method achieves performance comparable to state-of-the-art models while addressing the shortcomings of CLIP embeddings in handling outdoor scenes. By leveraging visual semantics directly, our method demonstrates enhanced robustness and adaptability in depth estimation tasks, showcasing its potential for application to other visual perception tasks.

\end{abstract}

\section{INTRODUCTION}
Estimating the distance between objects and the camera is crucial for many vision-based applications, including autonomous driving, virtual reality, robotic navigation and 3D reconstruction. Traditional depth sensing techniques often rely on specialized hardware like LiDAR and depth cameras (e.g., Kinect \cite{c1}), which can either be expensive or impractical for certain applications. Monocular depth estimation, which predicts depth from a single RGB image, offers a cost-effective and versatile alternative, eliminating the need for multiple cameras or complex hardware \cite{c2}\cite{c3}. However, the lack of geometric cues in 2D images makes monocular depth estimation an inherently ill-posed problem.

Deep learning has significantly advanced monocular depth estimation by learning complex mappings from images to depth maps. Eigen et al.\ pioneered this approach by introducing a convolutional neural network (CNN) that integrates global and local features \cite{c4}. Generative models have further enhanced depth estimation tasks by capturing more fine-grained details. For instance, Jung et al.\ employed generative adversarial networks (GANs) to produce more natural and detailed depth maps \cite{c5}. Despite their success, GANs face training difficulties such as mode collapse and convergence issues \cite{c6}, limiting their practicality.

\begin{figure}[thpb]
  \centering
  \includegraphics[width=\columnwidth]{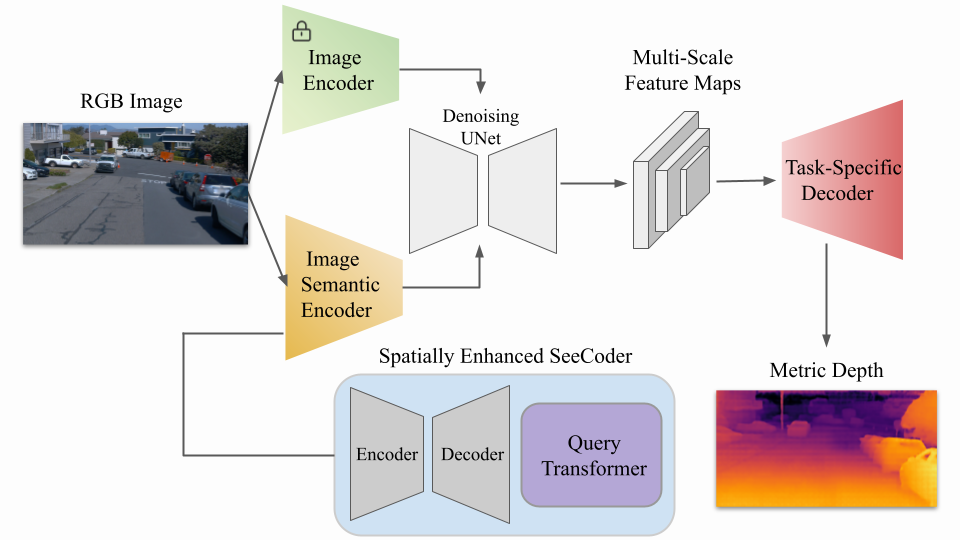}
  \caption{The overview of our proposed framework: an input RGB image is processed through two parallel pathways, an Image Encoder and an Image Semantic Encoder. The Image Encoder extracts latent features, with its weights frozen during training and inference, while the Image Semantic Encoder produces a semantic vector that conditions these features. Both are integrated within a denoising UNet that fuses these modalities to produce multi-scale feature maps. Subsequently, a decoder upsamples these feature maps, generating the final metric depth map.}
  \label{fig:architecture}
\end{figure}

Diffusion models have emerged as a superior alternative to GANs, offering training stability and capability to generate diverse samples \cite{c7}. In the context of depth estimation, diffusion models have shown promising results in monocular depth prediction tasks \cite{c8}\cite{c11}. However, models like Denoising Diffusion Probabilistic Models (DDPM) require extensive computational resources due to their iterative sampling processes \cite{c12}.

Stable Diffusion addresses this challenge by operating in a latent space rather than the high-dimensional pixel space, thereby reducing computational complexity while preserving image details \cite{c13}. In computer vision perception tasks, Zhao et al.\ proposed the Visual Perception with Diffusion (VPD) framework based on Stable Diffusion. By training a denoising UNet and a task-specific decoder, VPD achieved state-of-the-art results on the NYUv2 depth estimation dataset and the RefCOCO dataset \cite{c14}.

A critical component of Stable Diffusion's text-to-image capability is the Contrastive Language-Image Pre-training (CLIP) model, which provides rich semantic guidance by mapping textual descriptions to image features \cite{c15}. VPD leverages CLIP to incorporate semantic context into the model via text prompts. However, this approach is limited in complex environments, such as outdoor scenes, where generating accurate textual prompts is challenging due to the diversity and complexity of visual elements.

In order to overcome this limitation, we propose using SeeCoder, an alternative to CLIP, which directly extracts semantic context from images without relying on textual descriptions \cite{c16}. Additionally, we introduce a spatial enhancement module to further improve the extraction of visual features.

To validate the effectiveness and generalization ability of SeeCoder in depth estimation tasks, we integrated it into the VPD framework and conducted extensive experiments on KITTI \cite{c17} and three distinct scenarios from the Waymo Open Dataset \cite{c18}. Our method achieves performance comparable to state-of-the-art models, demonstrating its robustness in complex outdoor environments. The main contributions of our work can be summarized as follows:

\vspace{-0.5em}
\begin{itemize}
    \item We propose a novel approach that leverages SeeCoder to extract semantic context directly from images, addressing the limitations of text-based semantic guidance in depth estimation tasks.
    \item We incorporate a spatial enhanced module within SeeCoder to enhance performance on the KITTI dataset.
    \item We conduct extensive experiments on the KITTI and the Waymo Open Dataset, demonstrating that our method achieves performance competitive to state-of-the-art models.
\end{itemize}

\section{Related Work}
Monocular depth estimation is pivotal in applications such as robotic navigation, autonomous driving, and virtual reality. Traditional methods based on geometric principles, like Scale-Invariant Feature Transform (SIFT) and Conditional Random Fields (CRFs), often struggle with feature matching and computational demands in complex scenes. The advent of deep learning has significantly enhanced accuracy and efficiency, establishing learning-based approaches as the mainstream in monocular depth estimation \cite{c2}. Among these, generative models have shown substantial potential in depth prediction tasks \cite{c5}\cite{c8}\cite{c11}\cite{c19}. Below we discuss related work that have made use of generative models in depth estimation. 

\subsection{Generative Adversarial Networks}
Generative Adversarial Networks (GANs) \cite{c20} have been effectively applied to monocular depth estimation, particularly in enhancing fine-grained details. Jung et al. \cite{c5} utilized GAN-based generators to produce more detailed and realistic depth maps, improving depth estimation quality. Building upon this, Godard et al. introduced the MonoDepth2 model \cite{c19}, which employs a self-supervised learning framework for depth estimation. Instead of using a generator-discriminator architecture, MonoDepth2 relies on photometric consistency between consecutive frames or stereo pairs, enabling the model to learn depth without ground truth labels. The model introduces improvements such as minimum reprojection loss to handle occlusions and auto-masking to ignore invalid pixels, significantly advancing self-supervised depth estimation performance on datasets like KITTI. However, GANs often encounter training difficulties, such as mode collapse and convergence issues \cite{c6, c21}, which can limit their practical applicability. Additionally, GANs face challenges in generating high-resolution images, often struggling to maintain image quality and detail as the resolution increases \cite{c22}\cite{c23}.

\subsection{Diffusion Models}
Diffusion models have emerged as a promising alternative to GANs, offering improved training stability by iteratively adding and removing Gaussian noise through a probabilistic process modeled as a Markov Chain \cite{c12}. This approach leverages the gradual introduction of noise during the forward process and its removal during the reverse process, allowing for more stable training compared to GANs \cite{c7}. While they circumvent common GAN issues, diffusion models face challenges with slow sampling times due to their iterative nature. Nevertheless, they have achieved state-of-the-art results in depth estimation when trained on large datasets like Waymo and KITTI, notably with DepthGen \cite{c8}. DepthGen incorporates self-supervised pretraining, where the model learns general image structures from tasks like colorization, and multi-task training to enhance depth estimation accuracy. This combination allows the model to generalize well across different datasets. To address computational demands, Rombach et al. proposed applying diffusion processes in a lower-dimensional latent space, significantly enhancing efficiency without compromising accuracy \cite{c13}. Building on this approach, Zhao et al. introduced the Visual Perception with Diffusion (VPD) framework \cite{c14}, extending Stable Diffusion to various vision tasks. VPD integrates a denoising UNet and a task-specific decoder, achieving state-of-the-art performance on the NYUv2 depth estimation dataset and the RefCOCO image segmentation dataset, demonstrating robust results across diverse tasks.

\subsection{Semantic Embeddings in Diffusion Models}
Semantic encoders play a important role in providing contextual information for depth estimation within diffusion models. The Contrastive Language-Image Pre-training (CLIP) model \cite{c15} has been leveraged to supply rich semantic guidance, as seen in models like VPD, which achieved leading results on the NYUv2 dataset. However, reliance on text prompts can be limiting in complex scenes where generating accurate textual descriptions is challenging. To overcome this limitation, researchers have explored alternatives. BLIP-2 \cite{c24} generates image captions to serve as scene descriptions, aiding semantic understanding. Some other studies treat semantic extraction as a classification task, utilizing pretrained image encoders like Vision Transformer (ViT) \cite{c25} alongside learnable embedding vectors. For more direct and flexible applications, models such as SeeCoder offer a promising solution by extracting semantic features directly from visual inputs without relying on text-image alignment or classification tasks. Similar to CLIP, SeeCoder is trained on large datasets like Laion2B-en \cite{c26} and COYO-700M \cite{c27}, and its Transformer-based architecture enables it to capture rich, multi-scale semantic features, making it highly effective for various visual understanding tasks.

\subsection{Spatial Enhancement Techniques}
Enhancing spatial structure understanding is critical in depth estimation, especially for outdoor scenes where depth maps are usually sparse. Techniques like the Convolutional Block Attention Module (CBAM) \cite{c28} and dilated convolutions \cite{c29} have been proposed to improve spatial awareness. CBAM enhances feature representations by sequentially applying channel and spatial attention mechanisms, allowing the network to focus on informative features and relevant spatial locations. Meanwhile, dilated convolutions expand the receptive field without increasing the number of parameters, enabling the capture of multi-scale contextual information crucial for depth estimation. Both methods augment the model's ability to capture spatial dependencies without significantly increasing computational complexity, making them effective plug-and-play solutions for enhancing depth estimation accuracy.

\section{Methodology}

As depicted in Figure \ref{fig:architecture}, our proposed framework consists of four main components: a latent feature extractor, a spatially enhanced semantic encoder (SeeCoder), a denoising UNet, and a task-specific decoder. The input RGB image is processed through two parallel pathways: the latent feature extractor compresses the image into a lower-dimensional space, while the semantic encoder generates high-level embeddings for contextual information. These features are integrated within the denoising UNet to reconstruct the depth map by combining visual and semantic cues. Finally, the task-specific decoder refines and upsamples the output to produce the final high-resolution depth map.

\subsection{Latent Features Extraction}
In our pipeline, we utilize the encoder of a pre-trained variational autoencoder (VAE) to convert the RGB image from pixel space into a lower-dimensional latent representation, as is commonly done in Stable Diffusion \cite{c13}. Formally, the encoder maps the input image \( x \in \mathbb{R}^{H \times W \times 3} \) to a compact latent vector \( z \) via \( z = E_{\text{VAE}}(x) \), by learning a probabilistic distribution for each image. This process captures essential features while discarding irrelevant details, reducing computational cost and easing convergence. The VAE encoder remains frozen during training to retain its learned features, allowing our model to leverage these compact and informative representations for the subsequent depth map generation.

\subsection{Spatially Enhanced SeeCoder}
Unlike traditional methods that extract semantics using CLIP, our proposed spatially enhanced SeeCoder embedding directly extracts semantic context from the visual features of the image. The input image is processed by the SWIN-L \cite{c30} backbone encoder of SeeCoder, converting it into multi-scale feature maps. These feature maps are then upsampled through the SeeCoder decoder to obtain refined multi-scale representations. Finally, a Query Transformer performs semantic extraction, generating 148 semantic vectors of dimension 768. Among them, 144 are local queries, each performing cross-attention with the decoder's multi-scale feature maps to focus on specific regions, while 4 are global queries that concatenate with the local queries and perform self-attention to capture overall context \cite{c31}.

In order to enhance the model’s spatial understanding, we add dilated convolution and spatial attention modules to each transformer layer in the SeeCoder backbone encoder. The spatial attention module helps the model focus on important regions by generating an attention map based on the input features. Specifically, the attention map is calculated by applying average pooling and max pooling along the channel dimension, concatenating the results, and passing them through a convolution layer.

During training, we utilize the pretrained SeeCoder, freezing its weight parameters to retain the learned representations, and only update the weights for the newly added dilated convolution and spatial attention modules.

\subsection{Denoising UNet}

The denoising UNet is a pivotal component in our framework, responsible for reconstructing the depth map from the noisy latent representations generated during the diffusion process. The UNet architecture comprises an encoder and a decoder connected through skip connections, enabling the network to capture both global context and fine-grained details effectively.

In the encoder path, the UNet progressively downsamples the input latent features \( z_t \) at each time step \( t \), extracting hierarchical representations at multiple scales. Conversely, the decoder path upsamples these features to reconstruct the spatial dimensions, with skip connections merging corresponding encoder and decoder features. This design preserves essential information across different resolutions, facilitating accurate depth estimation.

In order to integrate semantic guidance, the denoising UNet incorporates cross-attention mechanisms that leverage the semantic embeddings \( s \) extracted by the spatially enhanced SeeCoder. At each resolution level, cross-attention layers enable the UNet to focus on relevant semantic information, enhancing the feature representations used for denoising. Specifically, the cross-attention operation is applied during both the upsampling and downsampling stages, allowing the model to align the latent features with the semantic context at multiple scales, improving the coherence and accuracy of the generated depth maps.

The UNet is trained to predict the noise \( \epsilon \) added to the latent representation \( z_0 \) during the forward diffusion process. The training objective minimizes the mean squared error between the predicted noise \( \epsilon_\theta(z_t, t, s) \) and the actual noise \( \epsilon \), formulated as:

\begin{equation}
L_{\text{DM}} = \mathbb{E}_{z_0, \epsilon \sim \mathcal{N}(0, 1), t} \left[ \left\| \epsilon - \epsilon_\theta\left(z_t, t, s \right) \right\|_2^2 \right],
\end{equation}

where \( z_t \) is obtained by the forward diffusion process defined as:

\begin{equation}
z_t = \sqrt{\bar{\alpha}_t} z_0 + \sqrt{1 - \bar{\alpha}_t} \epsilon,
\end{equation}

with \( \bar{\alpha}_t \) representing the accumulated product of the noise schedule parameters.

By optimizing this loss function, the denoising UNet learns to iteratively remove noise from the latent representations, progressively refining the depth map estimation. The incorporation of semantic embeddings via cross-attention enhances the denoising process, allowing the model to utilize high-level semantic cues alongside low-level visual features. This synergy between the latent features and semantic context enables our model to produce detailed and accurate metric depth maps, even in complex scenes.

During training, our denoising UNet is trained from scratch without relying on pre-trained weights. This allows the model to learn task-specific features tailored to monocular depth estimation directly from the data. By optimizing the entire network, the model learns to effectively utilize both the latent visual features and the high-level semantic cues to produce accurate depth maps. This end-to-end training approach ensures that all components of the model are finely tuned to the specific requirements of depth estimation, leading to better performance in complex scenes.

\subsection{Task-Specific Decoder}
In our task, the decoder serves to generate the final depth map. The decoder architecture comprises three primary components: a deconvolution module, a convolution module, and an upsampling module. It processes multi-scale feature maps as input, progressively refining and upsampling these features through deconvolution and convolution operations. For upsampling, we use bilinear interpolation with a scale factor of 2, ensuring smooth transitions between different resolutions. The final output is a high-resolution metric depth map, effectively capturing detailed spatial information from the input feature representations. The combination of deconvolution and bilinear upsampling enables efficient feature refinement and enhances spatial resolution without significantly increasing computational complexity.

\section{EXPERIMENTS}

\subsection{Outdoor Datasets}

\noindent \textbf{KITTI.} As a benchmark for autonomous driving and computer vision research, KITTI featuring real-world driving scenes from urban and rural areas. We use the data split of Eigen et al, specifically, using 23,158 annotated depth map and RGB image pairs for training, with another 697 RGBD image pairs reserved for validation.

\vspace{0.5em} \noindent \textbf{Waymo.} In order to conduct more comprehensive testing across different environments, we employed the Waymo Open Dataset. This dataset includes a vast collection of scenes from various cities, suburban areas, and rural roads across the United States, offering a broader range of weather conditions (e.g., sunny, rainy, foggy) and lighting conditions (e.g., daytime, dusk, nighttime). The Waymo Open Dataset provides high-resolution sensor data from multiple cameras and LiDAR, along with accurate ground truth annotations. This diversity allows us to evaluate our model's performance under varied and challenging scenarios, assessing its robustness and generalization capabilities in real-world conditions.

\subsection{Evaluation Metric}
In this study, three experiments were conducted to evaluate the effectiveness and generalization ability of the proposed method. The model was assessed using a quantitative analysis approach by comparing predicted depth maps to ground truth depth maps in the validation set. In the process of comparing differences, we only perform the calculations on valid depth values in the ground truth, which are the non-zero regions. To evaluate the proportion of predictions within different accuracy ranges, we employ the metrics $\delta_1$, $\delta_2$, and $\delta_3$, which represent the percentage of predictions where the relative error is within $1.25$, $1.25^2$, and $1.25^3$ times the actual depth values, respectively. Additionally, other metrics such as absolute relative error (Abs Rel), squared relative error (Sq Rel), and root mean square error (RMSE) are also utilized to provide a comprehensive assessment of the model's performance. Collectively, these metrics offer insights into both the accuracy and robustness of the depth predictions, ensuring a thorough analysis of the proposed method's effectiveness.

\subsection{Training and Evaluation on KITTI}

Our model was primarily trained on the KITTI dataset, where we trained various embedding schemes for approximately 20 epochs each. We recorded the best performance on the validation set for each scheme.

\vspace{0.5em} \noindent \textbf{Data Augmentation and Preprocessing.}
The preprocessing approach employed is consistent with that used in VPD. During training, various data augmentation techniques are applied to the RGB images from the KITTI dataset. These augmentations include random horizontal flipping, random cropping, adjustments to brightness and contrast, gamma correction, and random modifications to hue, saturation, and brightness. The goal of these augmentations is to enhance the model's robustness across different scenes and lighting conditions. During evaluation, the original resolution of 1242 x 375 depth maps are split into left and right segments for separate prediction. Each segment is then subjected to horizontal flipping augmentation. Finally, the two segments are weighted and fused to generate the final depth map.

\vspace{0.5em} \noindent \textbf{Implementation Details.}
Our model is implemented using PyTorch \cite{c32} and was trained end-to-end. We used AdamW as the optimizer with $\beta_0$ values of 0.9 and 0.999, a weight decay of 0.1. The model was trained on the KITTI dataset using eight NVIDIA L20 GPUs over approximately 13 hours, with a batch size of 3 per GPU, resulting in a total batch size of 24. We employed a one-cycle training strategy, starting with an initial learning rate of 4e-5, gradually increasing the learning rate to a maximum of 6e-4, and then decreasing it throughout the iterations.

\begin{figure}[thpb]
  \centering
  \includegraphics[width=\linewidth]{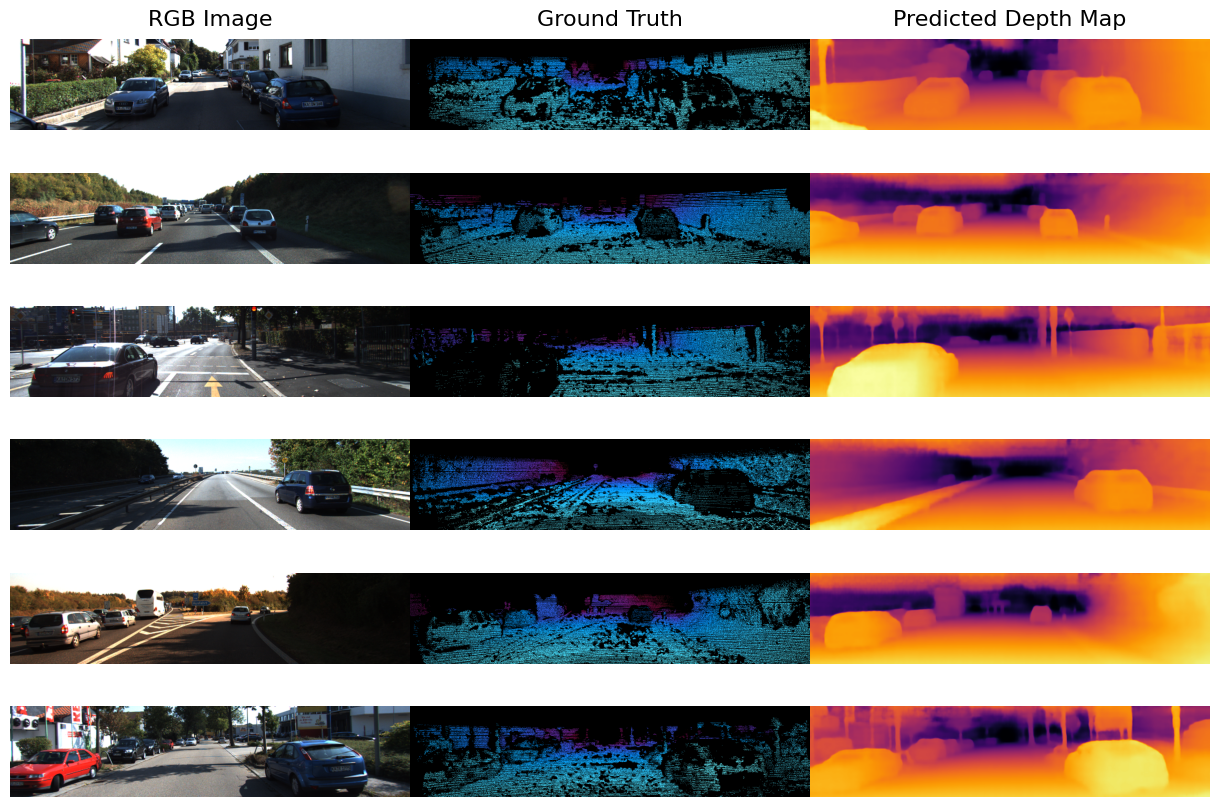}
  \caption{The visualization of selected samples from the KITTI dataset. Several samples are shown, with each column from left to right representing the RGB image, the sparse ground truth depth map after visualization processing, and the dense depth map predicted by our model.}
  \label{fig:kitti_fig}
\end{figure}

As shown in Table \ref{tab:kitti_ablations}, we evaluated four different embedding schemes. Using SeeCoder alone yielded satisfactory performance across all metrics. However, incorporating dilated convolutions (DC) or spatial attention (SA) individually led to increased errors, indicating training instability. Specifically, $\delta$ metrics declined, and RMSE, Abs Rel, and Sq Rel increased, suggesting less accurate distance predictions.

Combining both DC and SA with SeeCoder resulted in slight performance improvements. Notably, $\delta_1$ and $\delta_2$ metrics improved, indicating more accurate predictions within specified error thresholds. Although RMSE increased marginally, the overall accuracy was enhanced, demonstrating that the combined approach managed larger errors more effectively.

\begin{table}[h]
\begin{center}
\caption{Performance comparison of different embedding combination.}
\label{tab:kitti_ablations}
\resizebox{\columnwidth}{!}{
\begin{tabular}{lcccccc}
\toprule
\textbf{Embeddings} & $\delta_1 \uparrow$ & $\delta_2 \uparrow$ & $\delta_3 \uparrow$ & RMSE$\downarrow$ & Abs Rel$\downarrow$ & Sq Rel$\downarrow$  \\
\midrule
SeeCoder & 0.973 & 0.996 & 0.999 & 2.216 & 0.054 & 0.164 \\
SeeCoder+DC & 0.876 & 0.922 & 0.950 & 3.296 & 0.143 & 1.155 \\
SeeCoder+SA & 0.883 & 0.935 & 0.968 & 3.269 & 0.112 & 0.757 \\
\textbf{SeeCoder+DC+SA} & \textbf{0.974} & \textbf{0.997} & \textbf{0.999} & \textbf{2.179} & \textbf{0.052} & \textbf{0.162}\\
\bottomrule
\end{tabular}
}
\end{center}
\end{table}

\vspace{0.5em} \noindent \textbf{Ablation Study.}
We conducted an ablation study to assess the impact of spatial enhancement modules on model performance, using SeeCoder as the baseline. Integrating dilated convolutions (DC) alone decreased prediction accuracy, with $\delta_1$ dropping from 0.973 to 0.876 and Sq Rel increasing sixfold. Similarly, adding spatial attention (SA) alone did not improve performance significantly but was less detrimental than DC alone.

Interestingly, combining both DC and SA led to improved performance over the baseline. The $\delta_1$ metric increased to 0.974, and RMSE decreased from 2.216 to 2.179, indicating enhanced accuracy. These results suggest that the combined effect of DC and SA benefits the model more than either module individually. This may be because their combination compensates for each other's shortcomings; for instance, DC might introduce excessive contextual information, while SA may struggle to fully capture both global and local features.

\subsection{Evaluation on Waymo}

\noindent \textbf{Testing Details.}
During the testing process on the Waymo open dataset, we used the model weights that were trained on the KITTI dataset. We selectively chose depth maps from three distinct scene types for our testing. One of these scenes is a typical daytime outdoor scenarios similar to KITTI, serving as a reference for evaluating the model's generalization capability across different environments. The other two scenes are rainy and nighttime environments, each with nearly 600 test samples from the FRONT view. To maintain consistency with the KITTI training data, we first resized the resolution of 1920 x 1280 images proportionally. For RGB images, bilinear interpolation was used to smooth and preserve details, while for depth maps, nearest-neighbor interpolation was applied to maintain the accuracy of object edges. Consistent with the approach used for KITTI samples, we applied the Garg Crop \cite{c33} evaluation mask to minimize the influence of irrelevant information, such as the sky and the front of the vehicle, on the model's performance. Finally, performance metrics were calculated for each scene to observe the model's accuracy variation across different scenarios, and depth map visualizations were conducted to assess the model's performance.

\begin{figure}[thpb]
  \centering
  \includegraphics[width=\linewidth]{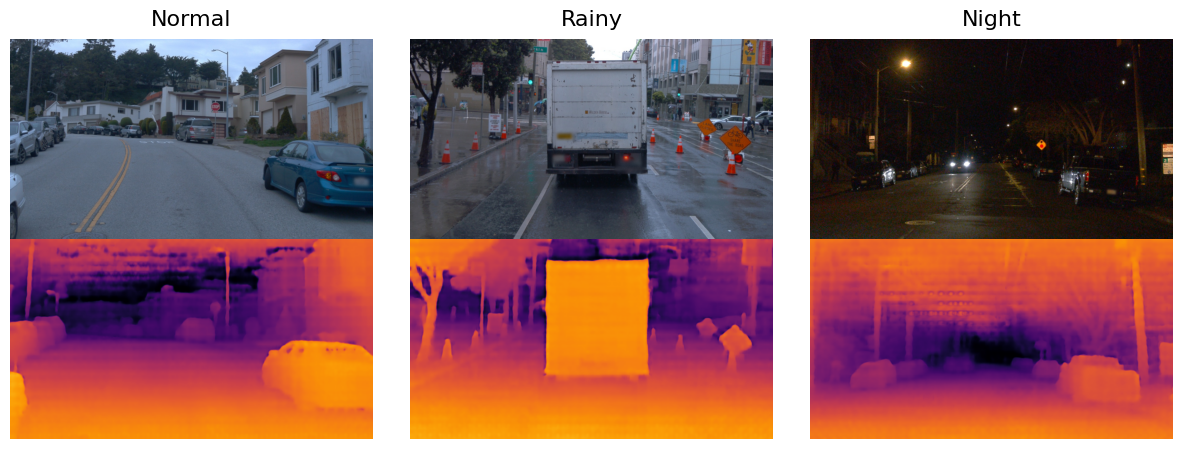}
  \caption{The visualization of model prediction differences across three different scenarios: normal daylight conditions, rainy weather, and nighttime.}
  \label{figurelabel}
\end{figure}

\vspace{0.5em} \noindent \textbf{Visualization Analysis.}
As illustrated in Figure 3, the predicted dense depth maps across the three different scenes accurately capture the edges and contours of objects. Even in areas where visual cues are less prominent, such as poorly lit regions in nighttime scenes, our method successfully identifies and completes objects. However, we also observed some potential challenges for accurate predictions. For instance, in Normal conditions, predictions of the sky and distant forests appear blurred, and the model struggles to precisely delineate the edges of these elements. This issue becomes more frequent in Rainy and Nighttime environments. In Rainy conditions, water droplets on the camera lens affect the accuracy of local predictions, while in nighttime, large dark regions without illumination lead to prediction failures. Such textureless, large color blocks pose increased difficulty for the model’s regional predictions, ultimately contributing to higher error rates.

\begin{table}[h]
\begin{center}
\caption{Performance comparison among different scenarios.}
\label{tab:waymo_scenes}
\resizebox{\columnwidth}{!}{
\begin{tabular}{lcccccc}
\toprule
\textbf{Scenes} & $\delta_1 \uparrow$ & $\delta_2 \uparrow$ & $\delta_3 \uparrow$ & RMSE$\downarrow$ & Abs Rel$\downarrow$  \\
\midrule
Normal & 0.044 & 0.312 & 0.893 & 5.061 & 0.389 \\
Rainy & 0.012 & 0.198 & 0.862 & 4.600 & 0.415 \\
Night & 0.017 & 0.150 & 0.807 & 5.678 & 0.428 \\
\bottomrule
\end{tabular}
}
\end{center}
\end{table}

As shown in Table \ref{tab:waymo_scenes}, testing on the Waymo dataset revealed a decline in performance metrics compared to the KITTI validation set, which is expected due to differences in data distribution caused by variations in the collection environment and sensor equipment. Despite this, the overall performance fluctuations across the three Waymo scenes were relatively small. For the global error metrics, RMSE and Abs Rel, the variation was around 10\%, indicating that our model demonstrates generalization and robustness across different scenes. However, we also observed significant fluctuations in the metrics $\delta_1$ and $\delta_2$ across different scenes. In rainy and nighttime conditions, these two metrics exhibited notably lower performance. This aligns with our visual observations, where extreme environments such as dark regions in nighttime scenes or raindrops in rainy conditions led to localized prediction failures, increasing the proportion of regional errors beyond the threshold.


\subsection{Comparison of related models}

In the field of monocular depth estimation, we compared our model's performance with several state-of-the-art methods on the KITTI Eigen split test set to comprehensively evaluate its effectiveness. The selected models include early CNN-based approaches, self-supervised learning methods, and recent transformer-based architectures. The metrics reported represent the best results cited in their respective papers. For example, Eigen et al. \cite{c4} introduced one of the pioneering CNN architectures for monocular depth estimation, establishing a baseline for subsequent research. Their method utilized a multi-scale network to predict depth. In \cite{c19}, Monodepth2 advanced the field with a self-supervised learning approach that leverages photometric consistency between stereo image pairs, significantly reducing the reliance on ground truth depth data and enhancing its practicality for real-world applications.

Recent transformer-based models have further pushed the boundaries of depth estimation performance. For example, Tu et. al \cite{c34}, introduced adaptive binning, allowing the model to allocate depth prediction resources more efficiently across different depth ranges. DPT \cite{c35} leveraged vision transformers to capture global context, enhancing the model's understanding of complex scenes. ZoeDepth \cite{c36}, particularly the ZoeDepth-M12-K version used in our comparison, focused on zero-shot generalization, demonstrating strong performance across various datasets without fine-tuning.

\begin{table}[h]
\begin{center}
\caption{Performance comparison of different models on KITTI validation set.}
\label{tab:compare_models}
\resizebox{\columnwidth}{!}{
\begin{tabular}{lcccccc}
\toprule
\textbf{Models} & \textbf{$\delta_1 \uparrow$} & \textbf{$\delta_2 \uparrow$} & \textbf{$\delta_3 \uparrow$} & RMSE$\downarrow$ & Abs Rel$\downarrow$ & Sq Rel$\downarrow$ \\
\midrule
Eigen et al.  & 0.702 & 0.898 & 0.967 & 6.307 & 0.203 & 1.517 \\
Monodepth2      & 0.879 & 0.961 & 0.982 & 4.701 & 0.115 & 0.882 \\ 
AdaBins        & 0.964 & 0.995 & 0.999 & 2.360 & 0.058 & 0.190 \\
DPT          & 0.959 & 0.995 & 0.996 & 2.573 & 0.062 & -     \\
ZoeDepth       & 0.970 & 0.996 & 0.999 & 2.440 & 0.054 & 0.189 \\
\midrule
\textbf{Ours}                 & \textbf{0.974} & \textbf{0.997} & \textbf{0.999} & 2.179 & \textbf{0.052} & \textbf{0.162} \\
\bottomrule
\end{tabular}
}
\end{center}
\end{table}

In the comparison with existing models, our model outperforms previous depth estimation methods across all evaluation metrics. This indicates that our approach, leveraging large-scale training and stable diffusion mechanisms, offers a significant advantage in capturing depth information from scenes. For instance, early CNN models had limitations in accuracy, while self-supervised models like Monodepth2, though reducing reliance on labeled data, still struggled to maintain high precision in certain complex scenarios. In contrast, recent transformer-based models such as AdaBins and DPT improved depth estimation by introducing adaptive binning and global context capturing. However, our model further enhances these performances, particularly in accurate estimation across different depth ranges and better global scene understanding, demonstrating its strong adaptability and generalization capabilities. By successfully repurposing SeeCoder for depth estimation, our model not only achieves near state-of-the-art performance on benchmark datasets but also shows potential for broader applications in future vision tasks.

\section{CONCLUSION AND FUTURE WORK}
We proposed a novel approach that integrates SeeCoder, an image-based semantic encoder, along with a spatial enhancement module into the Stable Diffusion framework for monocular depth estimation. By directly extracting semantic features from images and enhancing spatial features, our method addresses the limitations of text-based prompts in complex outdoor scenarios thereby improving depth estimation accuracy. Experiments on KITTI and Waymo datasets show that our approach achieves competitive performance when compared with state of the art techniques. Furthermore, spatially enhanced SeeCoder demonstrates robustness across conditions, including rain and night, suggesting that large-scale pre-trained models can be adapted to complex vision tasks. However, our model struggles with predicting depth in textureless regions and does not yet surpass current state-of-the-art models \cite{c11}. Future work includes improving SeeCoder's semantic accuracy beyond spatial enhancement, enhancing robustness to noise in textureless regions without cropping and extending our approach to other visual perception tasks to validate generalizability.

\section{Acknowledgement}
We would like to acknowledge the Engineering Physical Sciences Research Council (EPSRC) for their funding under the project EP/W014688/2, A digital COgnitive architecture to achieve Rapid Task programming and flEXibility in manufacturing robots through human demonstrations (DIGI-CORTEX).










\newpage
\begin{thebibliography}{99}

\bibitem{c1} J. Han, L. Shao, D. Xu, and J. Shotton, “Enhanced Computer Vision With Microsoft Kinect Sensor: A Review,” \textit{IEEE transactions on cybernetics }, vol. 43, no. 5, pp. 1318-1334, Oct. 2013.

\bibitem{c2} Y. Ming, X. Meng, C. Fan, and H. Yu, “Deep Learning for Monocular Depth Estimation: A Review,” \textit{Neurocomputing}, vol. 438, pp. 14-33, May 2021.

\bibitem{c3} C.S. Kumar, S.M. Bhandarkar, and M. Prasad. "Monocular depth prediction using generative adversarial networks." \textit{In Proceedings of the IEEE conference on computer vision and pattern recognition workshops}, pp. 300-308. 2018.

\bibitem{c4} D. Eigen, C. Puhrsch, and R. Fergus, “Depth Map Prediction from a Single Image Using a Multi-Scale Deep Network,” in \textit{Proceedings of the 27th International Conference on Neural Information Processing Systems (NIPS)}, vol. 2, pp. 2366–2374, Dec. 2014.

\bibitem{c5} H. Jung, Y. Kim, C. Oh, and K. Sohn, “Depth Prediction from a Single Image with Conditional Adversarial Networks,” in \textit{Proceedings of the IEEE International Conference on Image Processing (ICIP)}, Beijing, 2017, pp. 1717-1721.

\bibitem{c6} T. Salimans, I. Goodfellow, W. Zaremba, V. Cheung, A. Radford, and X. Chen, “Improved Techniques for Training GANs,” in \textit{Proceedings of the 30th International Conference on Neural Information Processing Systems (NIPS)}, pp. 2234–2242, Dec. 2016.

\bibitem{c7} P. Dhariwal and A. Nichol, “Diffusion Models Beat GANs on Image Synthesis,” in \textit{Proceedings of the 35th International Conference on Neural Information Processing Systems (NIPS)}, vol. 1, Article No. 672, pp. 8780–8794, Jun. 2024.

\bibitem{c8} S. Saxena, A. Kar, M. Norouzi, and D. J. Fleet, “Monocular Depth Estimation using Diffusion Models,” \textit{arXiv preprint}, arXiv:2302.14816, Feb. 2023.



\bibitem{c11} S. Patni, A. Agarwal, and C. Arora, “ECoDepth: Effective Conditioning of Diffusion Models for Monocular Depth Estimation,” in \textit{Proceedings of the IEEE/CVF Conference on Computer Vision and Pattern Recognition (CVPR)}, pp. 28285–28295, Jun. 2024.

\bibitem{c12} J. Ho, A. Jain, and P. Abbeel, “Denoising Diffusion Probabilistic Models,” in \textit{Proceedings of the 34th International Conference on Neural Information Processing Systems (NIPS)}, vol. 1, Article No. 574, pp. 6840–6851, Dec. 2020.

\bibitem{c13} R. Rombach, A. Blattmann, D. Lorenz, P. Esser, and B. Ommer, “High-Resolution Image Synthesis with Latent Diffusion Models,” in \textit{Proceedings of the IEEE/CVF Conference on Computer Vision and Pattern Recognition (CVPR)}, pp. 10684–10694, 2022.

\bibitem{c14} W. Zhao, R. Yongming, Z. Liu, B. Liu, J. Zhou, and J. Lu. "Unleashing text-to-image diffusion models for visual perception." \textit{In Proceedings of the IEEE/CVF International Conference on Computer Vision}, pp. 5729-5739. 2023.

\bibitem{c15} A. Radford, J.W. Kim, C. Hallacy, A. Ramesh, G. Goh, S. Agarwal, G. Sastry, A. Askell, P. Mishkin, J. Clark, and G. Krueger, "Learning transferable visual models from natural language supervision." \textit{In International conference on machine learning}, pp. 8748-8763. PMLR, 2021.

\bibitem{c16} X. Xu, J. Guo, Z. Wang, G. Huang, I. Essa, and H. Shi, "Prompt-free diffusion: Taking" text" out of text-to-image diffusion models." \textit{In Proceedings of the IEEE/CVF Conference on Computer Vision and Pattern Recognition}, pp. 8682-8692. 2024.

\bibitem{c17} A. Geiger, P. Lenz, C. Stiller, and R. Urtasun, “Vision Meets Robotics: The KITTI Dataset,” \textit{Int. J. Robot. Res.}, vol. 32, no. 11, pp. 1231-1237, Sep. 2013.

\bibitem{c18} P. Sun, H. Kretzschmar, X. Dotiwalla, A. Chouard, V. Patnaik, P. Tsui, J. Guo, Y. Zhou, Y. Chai, B. Caine and V. Vasudevan. "Scalability in perception for autonomous driving: Waymo open dataset." \textit{In Proceedings of the IEEE/CVF conference on computer vision and pattern recognition}, pp. 2446-2454. 2020.

\bibitem{c19} C. Godard, O. Mac Aodha, M. Firman, and G. J. Brostow, “Digging into Self-Supervised Monocular Depth Prediction,” in \textit{Proceedings of the IEEE/CVF International Conference on Computer Vision (ICCV)}, pp. 2019.

\bibitem{c20} I. Goodfellow, J. Pouget-Abadie, M. Mirza, B. Xu, D. Warde-Farley, S. Ozair, A. Courville, and Y. Bengio, "Generative Adversarial Nets," in \textit{Advances in Neural Information Processing Systems 27 (NIPS)}, 2014, pp. 2672–2680.

\bibitem{c21} M. Arjovsky, S. Chintala, and L. Bottou, “Wasserstein GAN,” \textit{arXiv preprint}, arXiv:1701.07875, Dec. 2017.

\bibitem{c22} A. Brock, J. Donahue, and K. Simonyan, “Large Scale GAN Training for High Fidelity Natural Image Synthesis,” in \textit{Proceedings of the International Conference on Learning Representations (ICLR)}, 2019.

\bibitem{c23} T. Karras, M. Aittala, S. Laine, E. Härkönen, J. Hellsten, J. Lehtinen, and T. Aila, “Alias-Free Generative Adversarial Networks,” in \textit{Proceedings of the 35th International Conference on Neural Information Processing Systems (NeurIPS)}, 2021.

\bibitem{c24} J. Li, D. Li, S. Savarese, and S. Hoi, “BLIP-2: Bootstrapping Language-Image Pre-training with Frozen Image Encoders and Large Language Models,” in \textit{Proceedings of the 40th International Conference on Machine Learning (ICML)}, vol. 2023, Article No. 814, pp. 19730–19742, Jul. 2023.

\bibitem{c25} A. Dosovitskiy, et al., “An Image is Worth 16x16 Words: Transformers for Image Recognition at Scale,” \textit{arXiv preprint}, arXiv:2010.11929, 2021.

\bibitem{c26} C. Schuhmann \textit{et al.}, “Laion-400m: Open Dataset of CLIP-filtered 400 Million Image-Text Pairs,” \textit{arXiv preprint}, arXiv:2111.02114, 2021.

\bibitem{c27} M. Byeon, B. Park, H. Kim, S. Lee, W. Baek, and S. Kim, “Coyo-700m: Image-text Pair Dataset,” 2022. https://github.com/kakaobrain/coyo-dataset. Accessed: Aug. 26, 2024.

\bibitem{c28} S. Woo, J. Park, J.-Y. Lee, and I. S. Kweon, “CBAM: Convolutional Block Attention Module,” in \textit{Proceedings of the 15th European Conference on Computer Vision (ECCV)}, Munich, Germany, Sep. 2018, pp. 3–19.

\bibitem{c29} F. Yu and V. Koltun, “Multi-Scale Context Aggregation by Dilated Convolutions,” in \textit{Proceedings of the International Conference on Learning Representations (ICLR)}, 2016.

\bibitem{c30} Z. Liu, Y. Lin, Y. Cao, H. Hu, Y. Wei, Z. Zhang, S. Lin, and B. Guo, "Swin Transformer: Hierarchical Vision Transformer Using Shifted Windows," in \textit{In Proceedings of the IEEE/CVF international conference on computer vision}, 2021, pp. 9992–10002.

\bibitem{c31} A. Vaswani, N. Shazeer, N. Parmar, J. Uszkoreit, L. Jones, A. N. Gomez, L. Kaiser, and I. Polosukhin, "Attention is All You Need," in \textit{Advances in Neural Information Processing Systems (NIPS)}, 2017, pp. 6000–6010.

\bibitem{c32} A. Paszke \textit{et al.}, “PyTorch: An Imperative Style, High-Performance Deep Learning Library,” in \textit{Advances in neural information processing systems(NeurIPS)}, 2019.

\bibitem{c33} R. Garg, B. G. V. Kumar, G. Carneiro, and I. Reid, "Unsupervised CNN for Single View Depth Estimation: Geometry to the Rescue," in \textit{Proc. Eur. Conf. Comput. Vis. (ECCV)}, 2016, pp. 740–756.

\bibitem{c34} Z. Tu, X. Chen, P. Ren, and Y. Wang, “AdaBin: Improving Binary Neural Networks with Adaptive Binary Sets,” in \textit{Proc. Eur. Conf. Comput. Vis. (ECCV)}, Cham, 2022, vol. 13671, pp. 379-395.

\bibitem{c35} R. Ranftl, A. Bochkovskiy and V. Koltun, "Vision transformers for dense prediction." \textit{In Proceedings of the IEEE/CVF international conference on computer vision}, pp. 12179-12188. 2021.

\bibitem{c36} S. F. Bhat, R. Birkl, D. Wofk, P. Wonka, and M. M{\"u}ller, “ZoeDepth: Zero-shot Transfer by Combining Relative and Metric Depth,” \textit{arXiv preprint}, arXiv:2302.12288, 2023.

\end{thebibliography}
\end{document}